\documentclass{article} 
\usepackage{conference_style,times}

\usepackage{amsmath,amsfonts,bm}

\def\eqref#1{equation~\ref{#1}}

\def\1{\bm{1}}

\DeclareMathAlphabet{\mathsfit}{\encodingdefault}{\sfdefault}{m}{sl}
\SetMathAlphabet{\mathsfit}{bold}{\encodingdefault}{\sfdefault}{bx}{n}

\usepackage{hyperref}
\usepackage{url}
\usepackage{graphicx}
\graphicspath{{resources/figures/}{figures/}{./}}
\usepackage{amsmath}
\usepackage{amssymb}
\usepackage{amsthm}
\usepackage{booktabs}
\usepackage{float}
\usepackage{xcolor}
\hypersetup{colorlinks=true, linkcolor=blue!55!black, citecolor=blue!55!black, urlcolor=blue!55!black}
\newtheorem{proposition}{Proposition}
\newtheorem{corollary}{Corollary}

\newcommand{\ci}[2]{{\scriptsize(#1, #2)}}
\newif\ifshowtodos \showtodostrue

\title{Wrong Votes, Right Gradients: Why Majority-Vote Rewards Track Verifiable RL}

\author{Dylan P. Wilson \\
Dartmouth College \\
\texttt{dylan.p.wilson.27@dartmouth.edu}
\And
Chengsong Huang \\
Washington University in St.\ Louis \\
\texttt{chengsong@wustl.edu} \\
\And
Jiaxin Huang \\
Washington University in St.\ Louis \\
\texttt{jiaxinh@wustl.edu} \\
}
\iclrfinalcopy 

\begin{document}

\maketitle
\fancyhead{}\lhead{}\chead{}\rhead{}

\vspace{-1.5em}
\begin{center}
  \small Code: \url{https://github.com/dylan518/wrong-votes-right-gradients}
\end{center}
\vspace{0.5em}

\begin{abstract}
Reinforcement learning with self-generated labels, such as majority vote over a model's own samples, often recovers most of the gain of training with verifiable rewards. This is puzzling from the usual prompt-level view, in which majority vote only returns a model's modal answer and cannot add information about correctness. We argue that the relevant unit is the batch-level update. Because parameters are shared, a prompt's own gradient accounts for less than 1\% of the change in that prompt's output at practical batch sizes, so a wrong label matters only to the extent that label errors share a direction across prompts. We decompose the majority-vote update into shrinkage and misdirection relative to the gold update, and show that misdirection vanishes with batch size whenever errors are incoherent across prompts, however often the labels are wrong. Measuring per-response gradients on five models, we find that majority-vote errors are as incoherent as a random-reward null, that the majority-vote update agrees with the gold update as closely as two independent gold updates agree with each other, and that its noise-corrected cosine with the gold update is 1 within error. Randomly flipping up to 30\% of gold labels leaves the update direction intact, while the same fraction flipped by a shared rule destroys it, and rewards that favor a kind of response on every prompt, such as a learned reward model, fall measurably short. The equivalence holds at every checkpoint we examine but is local: gold-reward training still ends ahead on held-out majority accuracy. Error coherence, not error rate, determines when self-generated labels can stand in for verification.
\end{abstract}

\section{Introduction}
\label{sec:intro}

Reinforcement learning is increasingly used to post-train language models, and its most reliable form uses verifiable rewards: programs, proof checkers, or reference answers that decide whether an output is correct \citep{shao2024deepseekmath,guo2025deepseekr1,zheng2026parallelr1}. Verification also confines training to existing labeled data or to domains with an automatic oracle \citep{zhao2026absolute}. Self-generated rewards remove this constraint by deriving feedback from the model's own outputs, such as agreement among sampled answers \citep{zuo2026ttrl,shafayat2025srt,huang2025rzero,yu2025rfew}, self-certainty \citep{zhao2026learning}, or entropy \citep{prabhudesai2025rent,agarwal2026unreasonable}. They often work surprisingly well: label-free training frequently recovers most of the improvement obtained with ground-truth rewards, and in some settings matches it \citep{zuo2026ttrl,shafayat2025srt,zhao2026learning,yu2026restrain} (Table~\ref{tab:related-recovery}). Yet the same rewards can also improve and then collapse \citep{shafayat2025srt,zhang2025nofreelunch,jin2026intrinsic}, or reinforce a model's existing preferences regardless of correctness \citep{he2026unsupervised}. When do self-generated labels work, and why?

The standard account is prompt-level \citep{he2026unsupervised,zhang2025consistent}. For a fixed prompt, majority vote (MV) converges to the model's modal answer, which need not be correct; it can sharpen the answer distribution but cannot extract correctness information that the distribution does not already contain. Several results sit uneasily with this account. TTRL raises the maj@64 of Qwen2.5-Math-7B on MATH-500 from 66.4 to 85.2, and its post-training average accuracy exceeds the initial maj@64 \citep[Fig.~7]{zuo2026ttrl}. In our own runs, held-out problems whose majority answer is wrong acquire a correct majority at about three times the rate that resampling alone produces (Section~\ref{sec:training}). Both effects require information to move between problems, which a prompt-level account does not allow.

We argue that the prompt is the wrong unit of analysis. Reinforcement learning sums reward-weighted gradients over a batch of diverse prompts and applies a single update to shared parameters. At practical batch sizes, a prompt's own gradient accounts for less than 1\% of the change in that prompt's output, so a wrong label barely steers its own prompt; what matters is whether the errors of many wrong labels add up to a common direction. We make this precise by decomposing the MV update into a component along the gold update, which only rescales the step, and an orthogonal component, misdirection, which changes its direction (Section~\ref{sec:cancel}). Misdirection is governed by the coherence of label errors across prompts rather than by their rate. Incoherent errors vanish once the batch exceeds a critical size set by the coherence of the gold signal, however often the labels are wrong, whereas errors that share a direction leave a floor that no batch size removes. That independent noise averages out is familiar from supervised learning \citep{rolnick2017labelnoise,chatterjee2020coherent}; what is new here is to apply it to the difference between two reward sources, to measure that difference directly in LLM post-training, and to turn it into a testable criterion for when self-generated labels can replace verification.

We test this account with per-response gradient sketches, which let us evaluate any reward on the same samples. On five models from two families, the MV update agrees with the gold update as closely as two independent gold updates agree with each other, and after correcting for prompt-sampling noise its cosine with the gold update is 1 within error (Section~\ref{sec:comparison}). MV errors are as incoherent as a random-reward null, while gold gradients share a direction that dominates beyond a few dozen prompts. Controls behave as the account predicts. Randomly flipping up to 30\% of gold labels leaves the update direction unchanged, whereas rewards that favor a kind of response on every prompt (frequent answers, confident answers, or those a learned reward model prefers) fall measurably short.

The equivalence is local. It holds at every checkpoint we examine, but it does not make the two training runs identical: gold training still ends ahead on held-out majority accuracy, and MV training mainly sharpens single-sample answers (Section~\ref{sec:training}). Our results therefore give a criterion rather than a guarantee. Self-generated labels can stand in for verification when a model's residual errors are sparse and specific to individual prompts, and they should fail where errors are shared, as with widely held misconceptions or prompts revisited until their errors accumulate.

\section{Preliminaries}
\label{sec:prelim}

\paragraph{Group-relative policy gradients.}
We train with GRPO \citep{shao2024deepseekmath}. A training step draws $B$ prompts $x_1,\dots,x_B$ and samples $n$ responses $y_{i1},\dots,y_{in}$ per prompt from the current policy $\pi_\theta$. A reward $r_{ij}\in\{0,1\}$ scores each response, and its advantage is the reward relative to its group, $A_{ij} = (r_{ij}-\bar r_i)/s_i$, where $\bar r_i$ is the group mean and $s_i$ is either the group standard deviation or $1$. At the sampling policy the importance ratios equal one and clipping is inactive, so the update direction is a sum of per-prompt gradients,
\[
G(r) = \sum_{i=1}^{B} g_i(r),
\qquad
g_i(r) = \sum_{j=1}^{n} A_{ij}\,\nabla_\theta \log \pi_\theta(y_{ij}\mid x_i).
\]
We call $g_i$ the prompt gradient and $G$ the batch update.

\paragraph{Gold and majority-vote rewards.}
Both rewards compare the final answer $a(y)$, taken from the last \verb|\boxed{}|, with a target. The gold reward uses the reference answer, $r^{\mathrm{gold}}_{ij} = \mathbf{1}[a(y_{ij}) = a^\ast_i]$. The majority-vote (MV) reward uses the most frequent answer in the group, $r^{\mathrm{mv}}_{ij} = \mathbf{1}[a(y_{ij}) = \hat a_i]$, and needs no label \citep{zuo2026ttrl,shafayat2025srt,huang2025rzero}. We write $G_{\mathrm{mv}}$ and $G_{\mathrm{gold}}$ for the two batch updates. When the majority answer is correct, $\hat a_i = a^\ast_i$, the two rewards coincide and so do the prompt gradients. The updates therefore differ only through prompts whose majority answer is wrong,
\[
E = G_{\mathrm{mv}} - G_{\mathrm{gold}} = \sum_{i:\,\hat a_i \neq a^\ast_i} \big(g_i(r^{\mathrm{mv}}) - g_i(r^{\mathrm{gold}})\big),
\]
which we call the MV error. A group in which every response receives the same \emph{gold} reward has zero gold advantage, so no cosine with gold is defined on it; we use the groups that are mixed under gold (164 of 256 prompts on Qwen3-4B-Base). The filter is on gold alone and forced rather than chosen: MV is often still live on the excluded groups.

\paragraph{Models and training.}
Our main model is Qwen3-4B-Base \citep{yang2025qwen3}, trained on DeepScaleR \citep{luo2025deepscaler} with $B=256$ prompts and $n=16$ responses per step. We train it twice from the same initialization, once with gold and once with MV rewards. Both runs see the same prompts at every step, so they differ only in the reward. Neither uses a KL or entropy term. For gradient analyses at initialization we add Qwen3-1.7B-Base, Qwen2.5-7B, Llama-3.2-3B-Instruct, and Llama-3.1-8B-Instruct \citep{grattafiori2024llama}. We evaluate on MATH-500 \citep{hendrycks2021math,lightman2024verify}, including its 134 level-5 problems, with 64 samples per problem at temperature 1. Appendix~\ref{app:setup} gives the full configuration.

\paragraph{Measuring updates.}
Comparing $G_{\mathrm{mv}}$ with $G_{\mathrm{gold}}$ requires both updates on the same responses. For every response we store the gradient of $\log\pi_\theta(y_{ij}\mid x_i)$ at the sampling policy, compressed by a count sketch \citep{charikar2002countsketch} to $K=2^{16}$ dimensions. Because $g_i$ is linear in the advantages, the update under any reward is a reweighting of the same per-response sketches, so every reward we compare is evaluated on identical samples. Cosines computed from sketches agree with exact fp32 gradients to within $0.004$. Our analyses use mean-centered advantages ($s_i=1$). Training divided by the group standard deviation; since $s_i$ is a per-group scalar, recomputing under that weighting moves $c_\infty$ by at most $0.010$ on any arm (Table~\ref{tab:stdnorm}).

\paragraph{Gold agreement as the benchmark.}
A cosine of one is the wrong benchmark: most of a prompt gradient is specific to that prompt, so even two gold updates on disjoint prompts are far from parallel. We therefore split the prompt groups into disjoint halves $\mathcal{A}$ and $\mathcal{B}$ and compute
\[
X = \cos\!\big(G^{\mathcal A}_{\mathrm{mv}}, G^{\mathcal B}_{\mathrm{gold}}\big),
\qquad
Y = \cos\!\big(G^{\mathcal A}_{\mathrm{gold}}, G^{\mathcal B}_{\mathrm{gold}}\big),
\qquad
Z = \cos\!\big(G^{\mathcal A}_{\mathrm{mv}}, G^{\mathcal B}_{\mathrm{mv}}\big).
\]
$Y$, the agreement of gold with itself, is the ceiling, and MV is as close to gold as the data can show when $X \approx Y$; pairing different halves keeps MV and gold from sharing sampling noise. If each half-batch update is its expectation plus independent prompt-sampling noise, the noise enters $X$, $Y$, and $Z$ only through their norms, so
\[
c_\infty = \frac{X}{\sqrt{YZ}}
\]
estimates the cosine between the expected MV and gold updates, which $\cos(G_{\mathrm{mv}}, G_{\mathrm{gold}})$ approaches as the batch grows. This is the classical correction for attenuation \citep{spearman1904proof}; as a ratio of estimates, $c_\infty$ can exceed 1 by sampling error, although the cosine it estimates cannot. We average over random splits, bootstrap over prompt groups, and report $c_\infty$ only when $Z \geq 0.3$, below which it is unstable.

\section{Gradient-Level Comparison of Majority Vote and Gold}
\label{sec:comparison}

We first ask how much the batch update changes when gold labels are replaced by MV. At initialization on all five models, it changes no more than prompt sampling already does; Section~\ref{sec:training} follows training.

\paragraph{MV agrees with gold as well as gold agrees with itself.}
Two gold updates on disjoint halves of the prompt groups agree only to a cosine of 0.37 to 0.84 (Table~\ref{tab:split-half}), because most of each prompt gradient is specific to its prompt. This is the ceiling for any label source, and MV reaches it: its cross-half cosine with gold is within 0.02 of the ceiling on four of the six arms and slightly above it on Qwen2.5-7B. The largest gap, on Llama-3.1-8B-Instruct (0.399 against 0.438), comes from MV's own noise (its split-half agreement is 0.362); after correcting for it, $c_\infty$ is 1.013. MV updates are as consistent with each other as gold updates are (Table~\ref{tab:split-half-full}).

\begin{table}[t]
\centering
\small
\caption{MV is as close to gold as gold is to itself. We split the prompt groups into two disjoint halves and compare summed gradients across the halves. \emph{Gold vs.\ gold} is the ceiling: how well two independent gold updates agree. \emph{MV vs.\ gold} pairs the MV update on one half with the gold update on the other; MV is indistinguishable from gold when the two columns match. $c_\infty$ removes the dependence on the number of prompt groups $N$ by estimating the cosine between the expected MV and gold updates; $c_\infty = 1$ means they point in the same direction. Table~\ref{tab:split-half-full} gives the full results.}
\label{tab:split-half}
\begin{tabular}{lrccc}
\toprule
 & & \multicolumn{2}{c}{Split-half cosine} & Expected update \\
\cmidrule(lr){3-4} \cmidrule(lr){5-5}
Model & $N$ & Gold vs.\ gold (ceiling) & MV vs.\ gold & $c_\infty$ (95\% CI) \\
\midrule
Qwen3-4B-Base$^\dagger$  & 331 & 0.647 & 0.647 & 1.006 \ci{.979}{1.023} \\
Qwen3-4B-Base            & 164 & 0.606 & 0.600 & 0.999 \ci{.965}{1.043} \\
Qwen3-1.7B-Base          & 636 & 0.842 & 0.834 & 0.998 \ci{.988}{1.006} \\
Qwen2.5-7B               & 151 & 0.799 & 0.818 & 1.010 \ci{.968}{1.021} \\
Llama-3.2-3B-Instruct    & 247 & 0.371 & 0.360 & 0.997 \ci{.921}{1.072} \\
Llama-3.1-8B-Instruct    & 208 & 0.438 & 0.399 & 1.013 \ci{.940}{1.073} \\
\bottomrule
\end{tabular}
\par\vspace{3pt}
\footnotesize
\raggedright
$^\dagger$Pools two batches of 256 prompts; the second was generated after one update. The row below is the on-policy batch alone.
Columns and $c_\infty$ use the same 200 partitions, so $c_\infty = X/\sqrt{YZ}$ checks row by row.
\end{table}

\begin{figure}[t]
\centering
\includegraphics[width=0.92\textwidth]{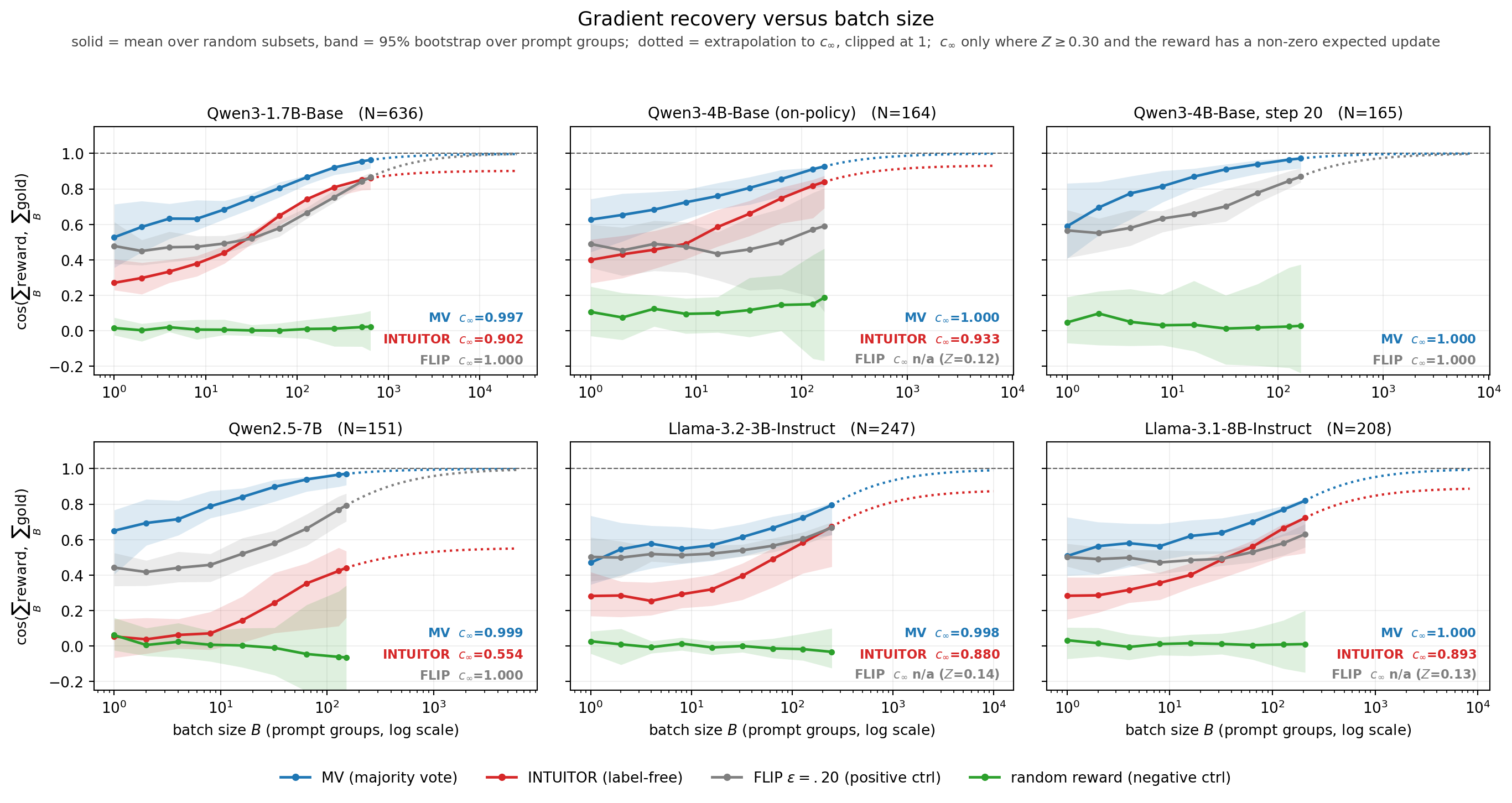}
\caption{Cosine between each reward's summed gradient and the summed gold gradient, against the
number of prompt groups $B$. Lines are means over random subsets; bands are 95\% bootstrap
intervals over prompt groups; dotted lines extrapolate to $c_\infty$. The flips (gray) and the
random reward (green) are averaged over 12 labellings, since a single frozen draw converges to its
own overlap with gold rather than its expectation. $c_\infty$ is clipped at 1 and shown only where
$Z\geq0.30$ and the expected update is non-zero; panels below that threshold are marked. INTUITOR
(red) is the raw reward (Table~\ref{tab:intuitor} gives the length-controlled variant).}
\label{fig:recovery}
\end{figure}

\paragraph{The expected MV update points in the gold direction.}
The ceiling depends on how many prompts each half contains; Figure~\ref{fig:recovery} and the last column of Table~\ref{tab:split-half} remove this dependence. The solid curves show the cosine between summed MV and gold gradients over random subsets of $B$ prompt groups. For a single prompt the two gradients coincide only when the majority is right, and the average cosine is 0.5 to 0.6. As $B$ grows the cosine rises toward its asymptote $c_\infty$, which is 1 within its 95\% interval on every model (point estimates 1.00 to 1.01), across two model families, base and instruction-tuned models, and 1.7B to 8B parameters. In expectation, the MV and gold updates point in the same direction; the finite-batch gap averages out, and Section~\ref{sec:cancel} shows why.

\paragraph{Each prompt moves with the batch.}
Why does a wrong label cost so little? Because it barely steers its own prompt. On Qwen3-1.7B-Base, a small step along a prompt's own gradient changes that prompt's output only 1.7 to 1.9 times as much as a step along another prompt's gradient (Table~\ref{tab:locality}). Summed over a batch, the other prompts dominate: at $B=64$ to $128$, a prompt's own gradient accounts for 0.09--0.38\% of the change in its output on the Qwen3 models, and 0.83\% on Llama-3.2-3B-Instruct at $B=64$ (Table~\ref{tab:prompt-share}).

The batch also sets the direction of that change. On the 76 prompts in the first Qwen3-4B-Base training batch whose majority answer is wrong, the MV batch update still increases the expected gold reward on 82\% of them, to first order. The prompt's own mislabeled term contributes $+0.007$ to the cosine between the update and the prompt's gold gradient; the other prompts contribute $+0.075$, almost exactly as much as under gold labels ($+0.074$). A wrong label is outvoted by the rest of the batch, so whether the MV update is right depends on how its errors add up across prompts.

\section{Why Label Errors Cancel}
\label{sec:cancel}

Section~\ref{sec:comparison} showed that the MV and gold updates point in the same direction in expectation. We now explain why, by separating two ways a label source can differ from gold and deriving when each survives a large batch.

\subsection{Shrinkage and misdirection}

Write the MV batch update relative to the gold update as
\[
G_{\mathrm{mv}} = (1-\beta)\, G_{\mathrm{gold}} + E_{\perp},
\qquad E_{\perp} \perp G_{\mathrm{gold}} .
\]
The coefficient $\beta$ measures \emph{shrinkage}, the part of the gold update that MV does not deliver, and $E_\perp$ measures \emph{misdirection}. With $R_\perp = \|E_\perp\| / \|G_{\mathrm{gold}}\|$,
\[
\cos\!\big(G_{\mathrm{mv}}, G_{\mathrm{gold}}\big)
= \frac{1-\beta}{\sqrt{(1-\beta)^2 + R_\perp^2}},
\qquad
\frac{\|G_{\mathrm{mv}}\|}{\|G_{\mathrm{gold}}\|}
= \sqrt{(1-\beta)^2 + R_\perp^2}.
\]
Only misdirection changes the direction of the update: under SGD, shrinkage acts like a smaller learning rate, and under an adaptive optimizer a roughly constant shrinkage is largely absorbed by the second-moment normalization. The question is why $R_\perp$ is small, and the answer lies in how vectors add.

\begin{proposition}[Growth of a sum]
\label{prop:growth}
For vectors $v_1,\dots,v_B$, let $s^2 = \frac1B\sum_i \|v_i\|^2$ and define the mean pairwise alignment
$\bar\rho = \sum_{i\neq j}\langle v_i, v_j\rangle \big/ \big((B-1)\sum_i\|v_i\|^2\big)$.
Then
\[
\Big\|\sum_{i=1}^B v_i\Big\|^2 = B\, s^2\,\big(1 + (B-1)\bar\rho\big),
\]
and the local growth exponent is
$\alpha(B) = \frac{d\log\|\sum_i v_i\|}{d\log B}
= \frac12 + \frac12\cdot\frac{B\bar\rho}{1+(B-1)\bar\rho}$.
\end{proposition}

Incoherent vectors ($\bar\rho = 0$) add like a random walk, growing as $B^{1/2}$, while vectors with a shared component grow as $B$ beyond the crossover $B^\ast \approx 1/\bar\rho$; a fitted exponent of $1/2$ therefore means no shared direction, and $\alpha \to 1$ a dominant one. For prompt gradients, $B^\ast$ is the gradient noise scale of \citet{mccandlish2018largebatch}: writing $g_i = \mu + \xi_i$ with independent noise $\xi_i$ gives $\bar\rho = \|\mu\|^2/(\|\mu\|^2 + \mathbb{E}\|\xi\|^2)$, so $B^\ast - 1 = \mathbb{E}\|\xi\|^2/\|\mu\|^2$. We apply the same accounting to the error between two rewards.

\begin{proposition}[Misdirection vanishes for incoherent errors]
\label{prop:misdirection}
Let a fraction $\varepsilon$ of the $B$ prompts carry a wrong majority label, and for each such prompt let $e_i$ be the component of $g_i(r^{\mathrm{mv}}) - g_i(r^{\mathrm{gold}})$ orthogonal to $G_{\mathrm{gold}}$. Let $s_e$ and $\rho_e$ be the root-mean-square norm and mean pairwise alignment of the $e_i$, and $s_g$ and $\rho_g$ those of the gold prompt gradients. Then
\[
R_\perp^2
= \varepsilon\,\frac{s_e^2}{s_g^2}\cdot
\frac{1 + (\varepsilon B - 1)\rho_e}{1 + (B-1)\rho_g}.
\]
If $\rho_e = 0$, then $R_\perp \to 0$ once $B \gg 1/\rho_g$, for any error rate $\varepsilon$. If $\rho_e > 0$, misdirection has a floor, $R_\perp \to \varepsilon\,(s_e/s_g)\sqrt{\rho_e/\rho_g}$, that no batch size removes.
\end{proposition}

The error rate $\varepsilon$ enters only as a prefactor; whether misdirection survives a large batch depends on the coherence $\rho_e$. Incoherent errors grow as $\sqrt{\varepsilon B}$ while the gold signal grows as $B$, so their ratio decays as $1/\sqrt{B\rho_g}$ however often the labels are wrong, whereas coherent errors grow as fast as the signal and leave a floor (proofs in Appendix~\ref{app:proofs}). For random label noise, the extreme case of incoherent error, the decomposition is exact in expectation.

\begin{corollary}[Independent label flips only shrink the update]
\label{cor:flips}
Suppose each gold reward is flipped independently with probability $\varepsilon < 1/2$, and advantages are mean-centered. Then the expected update given the responses is $(1-2\varepsilon)\,G_{\mathrm{gold}}$: the flips shrink the update by $\beta = 2\varepsilon$ and add zero-mean noise that is independent across prompts. The disattenuated cosine $c_\infty$ therefore equals 1 for every $\varepsilon < 1/2$.
\end{corollary}

These results predict that MV errors are incoherent, unlike gold gradients; that random flips preserve the update direction at any rate below one half; and that rewards whose errors recur across prompts fall short, however accurate they are.

\subsection{MV errors are incoherent}
\label{sec:incoherent}

Table~\ref{tab:coherence} reports the mean pairwise cosine between prompt gradients of Qwen3-4B-Base, with a Gaussian random reward as the null. At initialization, gold gradients share a direction (0.022 on prompts with a correct majority, 0.0066 on prompts with a wrong one), while the MV error $E$ is roughly 30 to 60 times less aligned (0.0004 and 0.0002), of the same order as the null. A single sketched cosine is accurate only to about 0.004, but each entry averages over thousands of prompt pairs, and the null, computed on the same sketches, absorbs any bias of the sketch.

\begin{table}[t]
\centering
\small
\setlength{\tabcolsep}{5pt}
\caption{MV errors are as incoherent as a random reward. Mean pairwise cosine between prompt gradients ($\times10^{-3}$); a positive value means the gradients share a direction. Gold gradients share one, while the MV error $E$ stays at the level of a Gaussian random reward (null). Qwen3-4B-Base along the MV run, split by whether the base model's majority answer is right or wrong. Because this majority is computed on separate samples, it can differ from the majority within a 16-response training group, so $E$ is not exactly zero on right-majority prompts. Checkpoints use different prompt draws; Table~\ref{tab:coherence-full} adds the MV gradients.}
\label{tab:coherence}
\begin{tabular}{lccccccc}
\toprule
 & \multicolumn{3}{c}{Right majority} & & \multicolumn{3}{c}{Wrong majority} \\
\cmidrule(lr){2-4} \cmidrule(lr){6-8}
Checkpoint & Gold & MV error $E$ & Null & & Gold & MV error $E$ & Null \\
\midrule
Base    & 22.41 & 0.39 & 0.30 & & \phantom{0}6.55 & 0.23 & 0.69 \\
Step 10 & \phantom{0}6.52 & 0.01 & 0.36 & & \phantom{0}0.82 & 0.15 & 0.50 \\
Step 20 & 27.99 & 0.14 & $-$0.15 & & 13.44 & 0.55 & 0.66 \\
\bottomrule
\end{tabular}
\par\vspace{3pt}
\footnotesize
\raggedright
The null is a Gaussian random reward on the same responses, taken from the gradients themselves rather than their difference with gold: a reward with zero expected update would otherwise inherit gold's coherence through $-G_{\mathrm{gold}}$ instead of providing a floor.
\end{table}

The growth of batch sums gives an independent estimate (Table~\ref{tab:exponents}). The error's exponent stays at $1/2$ at every scale, while the gold exponent rises from 0.58 at small batches to 0.87 at $B\approx150$, as Proposition~\ref{prop:growth} predicts for vectors with a shared component. The two estimates agree: inverting Proposition~\ref{prop:growth} at $B \approx 145$ gives $\bar\rho \approx 0.020$ for gold, against a measured 0.022, and a crossover $B^\ast \approx 50$, close to the $B^\ast = 37$ from a separate decomposition (Table~\ref{tab:prompt-share}). Practical batch sizes are well above this crossover.

\subsection{Error structure, not error rate}

Table~\ref{tab:controls} compares rewards whose errors differ in rate and structure, all on the same Qwen3-1.7B-Base responses. Independent flips isolate the error rate: as Corollary~\ref{cor:flips} predicts, $c_\infty$ stays at 1 for flip rates up to 30\%, so a high error rate alone does not misdirect the update.

Holding the rate fixed and changing only the structure separates the two. Flipping the \emph{same}
fraction by a rule shared across prompts, the longest correct and shortest incorrect responses,
leaves the rate unchanged but makes the errors coherent ($\rho_e$ rises from the null,
$0.2$--$1.4\times10^{-3}$, to $42$--$48\times10^{-3}$, above the gold gradients' own $7.9$). At a
matched $20\%$ error rate random flips give $c_\infty = 1.011$ and shared-rule flips $0.287$; at
$31\%$ the shared-rule update points away from gold ($-0.129$). The gap widens with $\varepsilon$
while the random-flip column does not move; Figure~\ref{fig:structure} shows the same contrast on all six arms.

The rewards that fall short favor a kind of response on every prompt. In expectation, random vote rewards each answer in proportion to its frequency, favoring frequent answers everywhere rather than only where the majority is wrong; self-certainty vote and INTUITOR favor confident answers; and an off-the-shelf reward model favors the responses it has learned to prefer, including confident ones \citep{leng2025taming}. Each preference is a direction shared across prompts, and each interval lies below 1 (random vote 0.961, self-certainty vote 0.975, length-controlled INTUITOR 0.969, reward model 0.982). A pure length reward, whose error has the same direction on every prompt, points away from gold ($c_\infty = -0.415$), and a reward with no signal stays near zero at every batch size (Figure~\ref{fig:recovery}). MV is the exception among the votes: it differs from gold only where the majority is wrong, and there its errors are specific to the prompt (Section~\ref{sec:incoherent}). Length-controlled INTUITOR stays at 0.94--0.97 on the Qwen models but falls to 0.76--0.79 on the instruction-tuned Llama models (Table~\ref{tab:intuitor}), although model family and instruction tuning are confounded here.

\begin{table}[t]
\centering
\small
\setlength{\tabcolsep}{5pt}
\caption{Error structure, not error rate, sets the update direction. Disattenuated cosine $c_\infty$ between the expected update of each reward and the expected gold update, all on the same Qwen3-1.7B-Base responses at initialization ($N=636$ prompt groups). A value of 1 means the reward moves the model in the gold direction; an interval below 1 means its errors add a systematic deviation. Wrong is the percentage of responses whose reward differs from gold; $\rho_e$ is the mean pairwise alignment of the per-prompt errors ($\times10^{-3}$). Random vote rewards a randomly chosen response's answer; self-certainty vote the answer with the highest aggregated self-certainty.}
\label{tab:controls}
\begin{tabular}{lccc}
\toprule
Reward & Wrong (\%) & $\rho_e$ ($\times10^{-3}$) & $c_\infty$ (95\% CI) \\
\midrule
\multicolumn{4}{l}{\emph{Independent noise: gold labels flipped at random with probability $\varepsilon$}} \\
\quad $\varepsilon = 0.05$ & \phantom{0}5.5 & \phantom{0}1.4 & 1.000 \ci{.979}{1.019} \\
\quad $\varepsilon = 0.10$ & \phantom{0}9.8 & \phantom{0}0.8 & 1.003 \ci{.977}{1.026} \\
\quad $\varepsilon = 0.20$ & 20.4 & \phantom{0}1.2 & 1.011 \ci{.969}{1.059} \\
\quad $\varepsilon = 0.30$ & 30.8 & \phantom{0}0.2 & 1.036$^{\ddagger}$ \\
\midrule
\multicolumn{4}{l}{\emph{Shared-rule noise: the same fraction flipped by response length}} \\
\quad $\varepsilon = 0.10$ & \phantom{0}9.7 & 48.3 & 0.773 \ci{.746}{.797} \\
\quad $\varepsilon = 0.20$ & 20.4 & 42.8 & 0.287 \ci{.249}{.323} \\
\quad $\varepsilon = 0.30$ & 31.4 & 42.0 & $-$0.129 \ci{$-$.160}{$-$.084} \\
\midrule
\multicolumn{4}{l}{\emph{Votes over sampled answers}} \\
\quad Majority vote        & \phantom{0}7.5 & \phantom{0}0.3 & 0.997 \ci{.989}{1.006} \\
\quad Self-certainty vote  & 23.3 & \phantom{0}1.6 & 0.975 \ci{.955}{.984} \\
\quad Random vote          & 18.9 & \phantom{0}6.3 & 0.961 \ci{.942}{.989} \\
\midrule
\multicolumn{4}{l}{\emph{Rewards that favor a kind of response on every prompt}} \\
\quad Reward model (GRM-Llama3.2-3B) & -- & 54.3 & 0.982 \ci{.974}{.991} \\
\quad INTUITOR, length-controlled    & -- & 41.6 & 0.969 \ci{.955}{.971} \\
\quad INTUITOR                       & -- & 48.5 & 0.902 \\
\quad Response length                & -- & 14.2 & $-$0.415 \ci{$-$.458}{$-$.399} \\
\bottomrule
\end{tabular}
\par\vspace{3pt}
\footnotesize
\raggedright
$^{\ddagger}$$Z = 0.27$ here, below the threshold for an interval. An error rate is given only for $\{0,1\}$ rewards; the four below are continuous scores, so
disagreement with gold is undefined for them. The $\rho_e$ null here is $1.1\times10^{-3}$: the independent flips scatter around it ($0.2$--$1.4$) and all keep $c_\infty$ at 1, while the rewards that fall short sit an order of magnitude higher. $\rho_e$ separates the regimes by scale, not by a threshold.

\subsection{Do the errors stay incoherent across steps?}
\label{sec:temporal}

Everything so far holds at fixed parameters. Extending it to a run needs the per-step errors to be
mutually incoherent too: to first order the gap between the two runs' parameters is
$\eta \sum_{t \le T} E_t$, which grows as $\sqrt{T}$ while the $E_t$ share no direction and as $T$
once they do. We sketch the MV run at steps $t = 3,\dots,11$ on each step's own on-policy rollouts
and score both rewards on identical samples.

The per-step errors are nearly, but not exactly, orthogonal. Over the thirty-six pairs $t \neq s$,
$\cos(E_t, E_s) = 0.030$, against $0.001$ for a random reward and $0.400$ for the gold gradients on
the same steps (Table~\ref{tab:temporal}); a sketched cosine resolves $0.004$, so the error carries
a small shared component, an order of magnitude below the signal's but above the floor. It shows in
the accumulation: $\|\sum_{t\le T} E_t\|$ grows with exponent $0.594$ against $0.762$ for gold, and
randomising the sign of each $E_t$ (preserving norms, destroying direction) gives a $\sqrt{T}$
reference of $0.540$ $[0.522, 0.576]$. The observed exponent lies above it ($p = 0.005$): the
error accumulates slightly faster than a random walk, while staying far short of the signal's
coherence.

The horizon is short, though the estimate is stable over it and the test strengthens steadily as it
grows: $p = 0.064$ over $t = 3,\dots,7$, then $0.026$, $0.013$, $0.009$ and $0.005$ as each further
step is added, with the exponent moving only between $0.594$ and $0.598$. Gradients at different $t$ sit
at different parameters, making these first-order comparisons.
\end{table}

\section{Training Dynamics}
\label{sec:training}

Sections~\ref{sec:comparison} and~\ref{sec:cancel} compare updates at fixed parameters, but training changes the model and with it the errors that MV makes. Following the gold and MV runs of Qwen3-4B-Base, which see identical prompts at every step, we ask whether the property behind the equivalence survives training, and whether equivalence at each step makes the two runs end in the same place.
Answer parsing degrades slowly rather than failing: the parseable fraction falls from $0.84$ at step 8 to $0.75$ at step 20 while the majority answer's accuracy stays flat, and what moves monotonically is the concentration of the vote (Appendix~\ref{app:setup}). We therefore report step 20 throughout.

\paragraph{The errors stay incoherent.}
MV errors stay at the random-reward null at steps 10 and 20, while gold gradients stay coherent (Table~\ref{tab:coherence}); at step 20 the error's growth exponent is still $1/2$ and the gold exponent reaches 0.90 (Table~\ref{tab:exponents}). The error keeps shrinking relative to the signal: $\|\sum E\|/\|\sum G_{\mathrm{gold}}\|$ falls from 0.43 at initialization to 0.24 at step 20, where $c_\infty = 1.000$ [0.972, 1.030]. By the decomposition in Table~\ref{tab:trajectory}, MV at step 20 delivers 98\% of the gold component and is rotated from the gold update by $14^\circ$. Step 10 is an outlier in every measure, including gold's agreement with itself.
Holding the prompts fixed confirms this: on the identical 256 step-21 prompts, shrinkage is flat ($\beta = 0.019 \to 0.020$) while misdirection halves ($R_\perp = 0.480 \to 0.239$), so the whole rise in batch cosine, $0.898 \to 0.972$, comes from misdirection (Table~\ref{tab:replication}). The step-10 excursion is not specific to one prompt draw: an independently seeded MV run gives the same depressed gold--gold ceiling at step 10 (0.276 against 0.248), with $c_\infty$ unreportable in both ($Z = 0.27$ and $0.25$). We do not have an explanation for it.

\begin{table}[t]
\centering
\small
\setlength{\tabcolsep}{5pt}
\caption{The MV update stays close to the gold update on the same batch. Qwen3-4B-Base along the MV run. Writing $G_{\mathrm{mv}} = (1-\beta)G_{\mathrm{gold}} + E_\perp$ (Section~\ref{sec:cancel}), shrinkage $\beta$ is the fraction of the gold step that MV does not deliver (negative: MV moves further along the gold direction than gold does), and misdirection $R_\perp = \|E_\perp\|/\|G_{\mathrm{gold}}\|$ is the size of the part pointing elsewhere. Cosine is $\cos(G_{\mathrm{mv}}, G_{\mathrm{gold}})$ on the same batch, error/signal is $\|\sum E\|/\|\sum G_{\mathrm{gold}}\|$, and ceiling is the split-half gold vs.\ gold cosine, which is anomalously low at step 10.}
\label{tab:trajectory}
\begin{tabular}{lccccc}
\toprule
Checkpoint & Cosine & Shrinkage $\beta$ & Misdirection $R_\perp$ & Error/signal & Ceiling \\
\midrule
Base    & 0.927 & $-$0.055 & 0.428 & 0.432 & 0.647$^\dagger$ \\
Step 10 & 0.810 & \phantom{$-$}0.143 & 0.621 & 0.637 & 0.265 \\
Step 20 & 0.972 & \phantom{$-$}0.020 & 0.239 & 0.240 & 0.658 \\
\bottomrule
\end{tabular}
\par\vspace{3pt}
\footnotesize
\raggedright
$^\dagger$The pooled arm of Table~\ref{tab:split-half}.
\end{table}

\paragraph{The runs track each other on the training distribution.}
Evaluated on both runs, the MV reward rises almost as fast in the gold run as in the run that optimizes it (Figure~\ref{fig:mv-reward-gold}, Appendix~\ref{app:additional}), because the two rewards coincide wherever the majority is correct. Training accuracy is indistinguishable through step 15 (prompt-paired difference $+0.002$ [$-0.001$, $+0.006$] over 3,840 pairs). The runs differ in the weight placed on wrong majorities: the gap between MV reward and accuracy grows from 0.09 to 0.15 in the MV run but stays at 0.09 to 0.10 in the gold run.

\paragraph{MV repairs wrong majorities through other prompts.}
Of the 134 level-5 MATH-500 problems, 66 have a wrong majority under the base model; by step 20 of the MV run, 13 of them (19.7\%) have a correct one (Table~\ref{tab:maj-flips}), about three times the flip rate between two independent halves of the samples. Since the run does not train on these problems, the repair comes from the shared update. Breakage runs at 4.4\%, below the resampling rate. The gains are largest at small $k$, where maj@1 and maj@4 rise by 17 to 19 points, but maj@16 and maj@64 rise significantly too ($+0.100$ and $+0.082$). MV training changes the answer the model commits to on a single sample more than it changes the vote.
The evaluation problems are all but disjoint from training: against the 29k DeepScaleR problems, one level-5 problem appears verbatim and two more exceed a $0.80$ token-overlap threshold (Appendix~\ref{app:setup}). Dropping all three leaves the repair rate inside its interval.

\paragraph{Local equivalence does not make the trajectories identical.}
On held-out problems, gold training ends ahead. At step 10 on the level-5 problems, averaging over two seeds per reward, gold leads MV in maj@1 (0.344 vs.\ 0.320) and maj@64 (0.590 vs.\ 0.541), though not in pass@64 (0.855 vs.\ 0.862); on all of MATH-500 at step 20, maj@64 rises from 0.747 to 0.772 with MV and to 0.845 with gold. Doubling the MV batch to 512 prompts does not close the gap (step-10 maj@64 $0.545$ vs.\ $0.582$, paired 95\% CI $[-0.007, +0.086]$), as expected when 256 prompts is already far above $B^\ast$, so the gap must come from something that per-step misdirection does not capture. Labels must also come from the current policy: labels from the frozen base model are nearly as accurate as on-policy MV but deliver far less of the gold update (Appendix~\ref{app:additional}).
Paired over the 500 problems, MV's gain over base is significant at every $k$ but collapses with it, from $+0.237$ [$+0.223$, $+0.252$] at maj@1 to $+0.026$ [$+0.005$, $+0.047$] at maj@64, against $+0.262$ and $+0.099$ for gold. At step 10 gold's advantage is weaker than the spread between seeds (Table~\ref{tab:seedband}).

The property behind the equivalence thus persists through training, yet the runs separate on held-out majority accuracy: equivalence at each step bounds the error an update adds, not how errors accumulate (Section~\ref{sec:conclusion}).

\section{Related Work}
\label{sec:related}

\paragraph{Self-generated rewards.}
Models can improve from their own outputs, by fine-tuning on self-consistent answers \citep{huang-etal-2023-large,wang2023selfconsistency} or by judging their own responses \citep{yuan2024selfrewarding,li2026selfrewarding}; see \citet{tao2024survey} and \citet{tan2024large} for surveys. In RL for reasoning, the reward can come from majority vote \citep{zuo2026ttrl,shafayat2025srt}, from confidence or entropy \citep{kang2025selfcertainty,zhao2026learning,huang2026selfcalibration,prabhudesai2025rent,agarwal2026unreasonable,zhang2025empo}, or from stabilized variants \citep{zhou2025evolrl,zhang2025corewarding,yu2026restrain}. Self-play systems also generate the tasks themselves \citep{huang2025rzero,yu2025rfew,chen2025sqlm,huang2026g,huang2026skill,yang2026ttcs,zhao2026absolute,liu2026spiral,liu2025spice,yue2026dr,shi2026learning,huang2026envharness,he2025visplay,li2026mm,li2026structured}, sometimes with checks on the generated data \citep{lai2026verifier,shen2026let,liang2026mathdebuggerdetectingdiagnosingerrors}. Rather than propose a new reward, we explain when such rewards can replace verification.

\paragraph{Label-free reinforcement learning.}
When label-free training works remains unsettled \citep{rahman2026weak,he2026unsupervised}. It can collapse \citep{shafayat2025srt,zhang2025nofreelunch,jin2026intrinsic}, reinforce the model's own preferences \citep{he2026unsupervised,arazo2020confirmation}, or lose the correct answer when the consensus is wrong \citep{lin2026extinction}, which has motivated complementary signals and verifiers \citep{yan2026scrl,pan2026coverrl,wang2025beyondmv,liao2026toolverification}. Gains can also come from spurious rewards or contamination \citep{shao2025spurious,chen2025exploration,wu2025memorization}, and RL may mostly sharpen existing behavior unless exploration is broadened \citep{yue2025doesrl,huang2026nonsense}. Existing explanations are prompt-level \citep{he2026unsupervised,zhang2025consistent}. Ours is batch-level: in the spirit of label-noise robustness \citep{rolnick2017labelnoise} and shared gradient directions \citep{chatterjee2020coherent,mccandlish2018largebatch,wei2026relex}, it asks whether the error between two reward sources has a shared component.

\section{Conclusion and Limitations}
\label{sec:conclusion}

We have argued that the batch-level update, not the individual prompt, is the right unit for understanding self-generated rewards. A fixed policy can have a wrong modal answer on a given problem, and MV will return it; but reinforcement learning updates shared parameters from many diverse prompts, and a wrong label matters only when errors across a batch share a direction. In our measurements they do not: a pretrained model's residual errors are sparse and prompt-specific, stay incoherent across prompts at every checkpoint we measure, and cancel in aggregate, while structure shared across prompts is reinforced.

The same view clarifies what verification adds. With any reward, GRPO improves only behavior that can be expressed as a coherent change to one shared policy, so the advantage of verifiable over self-generated labels should be largest for errors that recur across many prompts: in domains with widely shared misconceptions \citep{lin2022truthfulqa,prelec2017surprisingly}, or when the same prompts are revisited until their errors accumulate. For label-free methods, the criterion suggests that making errors incoherent, by labelling with the current policy (Appendix~\ref{app:additional}) and drawing fresh and diverse prompts, matters as much as making them rare.

\paragraph{Limitations.}
Our training runs use one model, Qwen3-4B-Base, and Qwen models benefit even from rewards unrelated to correctness \citep{shao2025spurious}; only the gradient-level results cover five models from two families. The runs are short and unregularized, and beyond step 15 the MV run's entropy rises sharply, a regime we do not analyze. We have two runs per reward and measure cosines of raw gradients, whereas Adam's preconditioner can reweight directions. Most importantly, per-step equivalence does not establish equivalence of trajectories, and over the nine steps we measure the errors already accumulate slightly faster than a random walk (Section~\ref{sec:temporal}).


\section*{Reproducibility Statement}
Appendix~\ref{app:setup} lists the training configuration, prompt format, evaluation protocol, and gradient-sketch procedure, and Appendix~\ref{app:proofs} gives all proofs.
We will release, under an open licence: the gradient-sketching code and the count-sketch implementation; the reward-reweighting, split-half and disattenuation estimators that produce every number in this paper; the scripts that generate each table and figure; and the launcher and configuration files for every training and evaluation run, including the majority-vote advantage estimator. We will also release the per-response gradient sketches, the gold and majority-vote label sets, the full text of every generation used in the analyses rather than extracted answers alone, and the dated experiment log recording each run's configuration, failures, and diagnosed causes, so that both the reported analyses and the path taken to them can be reproduced.

\bibliography{refs}
\bibliographystyle{conference_style}

\appendix
\section{Experimental Details}
\label{app:setup}

\paragraph{Training.}
Table~\ref{tab:hparams} lists the configuration shared by the gold and MV runs. The two runs start from the same checkpoint and use the same data order, so every step trains on the same prompts. Each step processes 256 prompts in two optimizer updates of 128 prompts. The training loss uses std-normalized advantages for both rewards.

\begin{table}[H]
\centering
\small
\caption{Training configuration shared by the gold and MV runs.}
\label{tab:hparams}
\begin{tabular}{ll}
\toprule
Setting & Value \\
\midrule
\multicolumn{2}{l}{\emph{Model and data}} \\
\quad Base model & Qwen3-4B-Base \\
\quad Training data & DeepScaleR (29k problems) \\
\quad Framework & verl 0.8.0 \citep{sheng2025hybridflow} \\
\quad Training steps & 26 \\
\midrule
\multicolumn{2}{l}{\emph{Batch}} \\
\quad Prompts per step ($B$) & 256 \\
\quad Responses per prompt ($n$) & 16 \\
\quad Mini-batch & 128 prompts (2 updates per step) \\
\quad PPO epochs & 1 \\
\midrule
\multicolumn{2}{l}{\emph{Optimization}} \\
\quad Learning rate & $1\times10^{-6}$, constant \\
\quad Clip range & $\epsilon_{\text{low}}=0.2$, $\epsilon_{\text{high}}=0.28$ \\
\quad Loss aggregation & token mean \\
\quad Advantage & group mean-centered, divided by group std \\
\quad KL / entropy coefficient & 0 / 0 \\
\quad Gradient clipping & 1.0 \\
\quad Weight decay & 0.01 \\
\midrule
\multicolumn{2}{l}{\emph{Sampling}} \\
\quad Temperature, top-$p$, top-$k$ & 1.0, 1.0, 40 \\
\quad Max response length & 4096 tokens \\
\bottomrule
\end{tabular}
\end{table}

\paragraph{Prompt format and answer extraction.}
Prompts use the Qwen chat template with the system prompt ``Please reason step by step, and put your final answer within \verb|\boxed{}|.'' Thinking mode is disabled, so the response begins with an empty thinking block. The answer of a response is the content of the last \verb|\boxed{}| in its final 400 characters.
A response whose answer cannot be parsed takes no part in the vote and receives reward 0; ties are broken toward the answer that appears first in the group. The training reward and the reward applied to the stored sketches use the same rule.

\paragraph{Evaluation.}
We evaluate on MATH-500 and on its 134 level-5 problems, drawing 64 samples per problem at temperature 1.0 and top-$p$ 1.0 with a 4096-token limit. maj@$k$ is the accuracy of the most frequent parsable answer among $k$ samples.

\begin{table}[H]
\centering
\small
\caption{Same-prompt replication (Section~\ref{sec:training}). Base and step 20 evaluated on the
\emph{identical} 256 step-21 prompts, each scored on its own on-policy rollouts. Shrinkage is
unchanged while misdirection halves, and the two reproduce the measured batch cosine. Steps 5 and
15 are absent because the run does not checkpoint there.}
\label{tab:replication}
\begin{tabular}{lcc}
\toprule
 & Base & Step 20 \\
\midrule
Mixed-outcome prompts (of 256)                & 125 & 165 \\
\quad wrong- / right-majority                  & 41 / 84 & 81 / 84 \\
\midrule
$\cos(G_{\mathrm{mv}}, G_{\mathrm{gold}})$    & 0.898 & \textbf{0.972} \\
$\|G_{\mathrm{mv}}\|/\|G_{\mathrm{gold}}\|$   & 1.092 & 1.009 \\
Shrinkage $\beta$                             & $+0.019$ & $+0.020$ \\
Misdirection $R_\perp$                        & 0.480 & \textbf{0.239} \\
$(1-\beta)/\sqrt{(1-\beta)^2+R_\perp^2}$      & 0.898 & 0.972 \\
\bottomrule
\end{tabular}
\end{table}

\begin{table}[H]
\centering
\small
\caption{Std-normalized advantages give the same answer. $c_\infty$ between the MV and gold updates
with advantages mean-centered ($s_i = 1$) and with $s_i$ set to the group standard deviation, the
weighting the training loss uses, computed from the same stored sketches. Step 10 is omitted: its
own split-half agreement is $Z \approx 0.25$, below the threshold at which $c_\infty$ is reported.}
\label{tab:stdnorm}
\begin{tabular}{lrcc}
\toprule
Model & $N$ & Mean-centered & Std-normalized \\
\midrule
Qwen3-1.7B-Base            & 636 & 0.997 \ci{.989}{1.006} & 0.994 \ci{.982}{1.008} \\
Qwen3-4B-Base (on-policy)  & 164 & 0.999 \ci{.965}{1.043} & 1.000 \ci{.968}{1.034} \\
Qwen3-4B-Base step 20      & 165 & 1.002 \ci{.981}{1.015} & 1.002 \ci{.985}{1.013} \\
Qwen2.5-7B                 & 151 & 0.998 \ci{.968}{1.027} & 0.996 \ci{.962}{1.030} \\
Llama-3.2-3B-Instruct      & 247 & 0.996 \ci{.929}{1.078} & 0.998 \ci{.911}{1.096} \\
Llama-3.1-8B-Instruct      & 208 & 1.007 \ci{.944}{1.079} & 1.017 \ci{.929}{1.119} \\
\bottomrule
\end{tabular}
\end{table}

\begin{figure}[H]
\centering
\includegraphics[width=\linewidth]{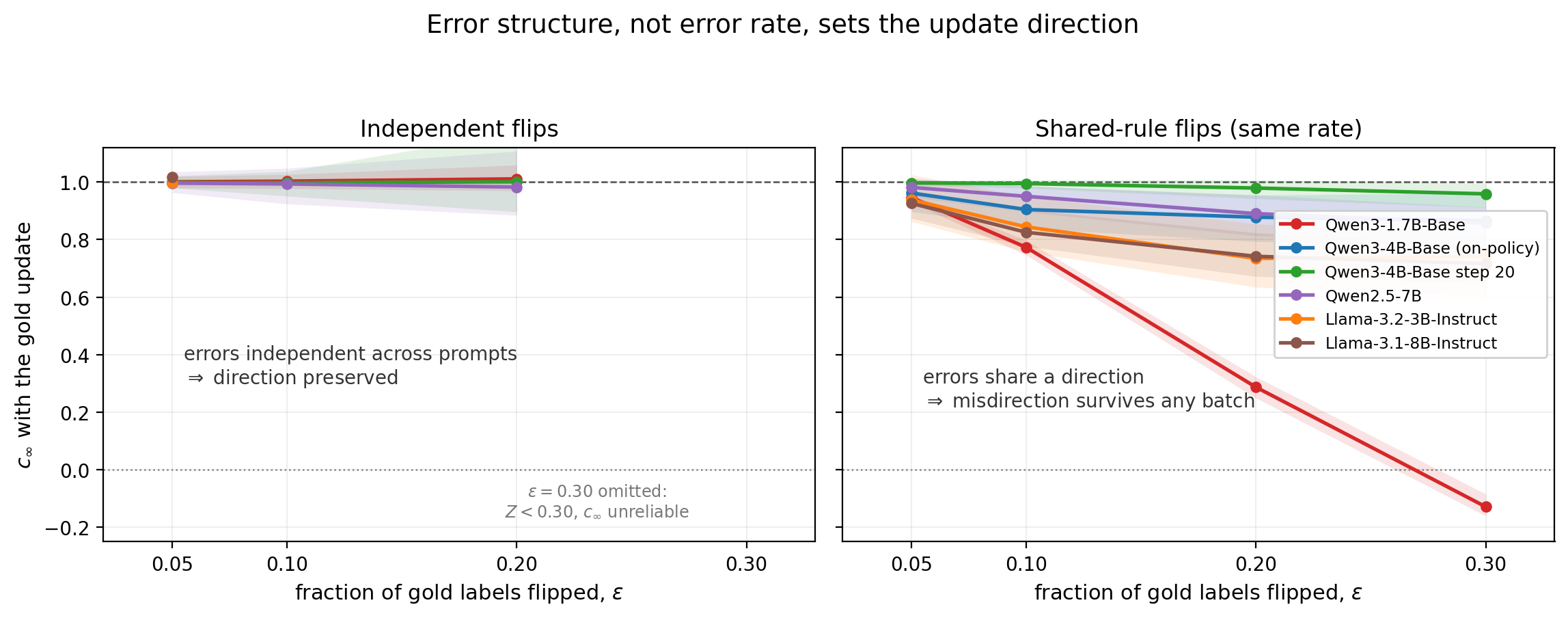}
\caption{Error structure, not error rate, sets the update direction (Section~\ref{sec:cancel}).
Both panels corrupt the same fraction $\varepsilon$ of gold labels on the same responses and differ
only in how the flips are chosen: independently per response (left), or by a rule shared across
prompts, the longest correct and shortest incorrect responses (right). Independent flips leave
$c_\infty$ at 1 on every model; shared-rule flips fall monotonically on every model, and on
Qwen3-1.7B-Base the update turns away from gold entirely by $\varepsilon = 0.30$. Bands are 95\%
bootstrap intervals. We plot $c_\infty$ rather than the raw cosine against batch size because the
two schemes differ in their own split-half agreement $Z$ and so are attenuated by different
amounts; the raw curve can rank the shared-rule flip above the independent one, which reflects
its lower sampling noise rather than better alignment. Points with $Z < 0.30$ are omitted.}
\label{fig:structure}
\end{figure}

\begin{table}[H]
\centering
\small
\caption{Run-to-run variation is comparable to the gold--MV gap (Section~\ref{sec:training}). Two
MV runs and two gold runs, trained from different seeds, all evaluated at step 10 on the same 134
level-5 problems with 64 samples each. maj@$k$ for $k<64$ averages over random $k$-subsets drawn
without replacement; at $k=64$ it uses the whole pool. Answers are clustered and graded exactly as
the evaluation harness does, including its treatment of ties, so these numbers are directly
comparable with those quoted in Section~\ref{sec:training}.}
\label{tab:seedband}
\begin{tabular}{lcccc}
\toprule
Run & maj@1 & maj@16 & maj@64 & pass@64 \\
\midrule
MV, seed A    & 0.325 & 0.525 & 0.545 & 0.858 \\
MV, seed B    & 0.314 & 0.505 & 0.537 & 0.866 \\
Gold, seed A  & 0.348 & 0.550 & 0.585 & 0.843 \\
Gold, seed B  & 0.339 & 0.557 & 0.595 & 0.866 \\
\midrule
\emph{Within-reward spread} (MV)   & 0.011 & 0.019 & 0.007 & 0.007 \\
\emph{Within-reward spread} (gold) & 0.008 & 0.007 & 0.010 & 0.022 \\
\emph{Gold $-$ MV}, mean of seeds  & $+$0.024 & $+$0.039 & $+$0.049 & $-$0.007 \\
\bottomrule
\end{tabular}
\end{table}

\begin{figure}[H]
\centering
\includegraphics[width=0.62\linewidth]{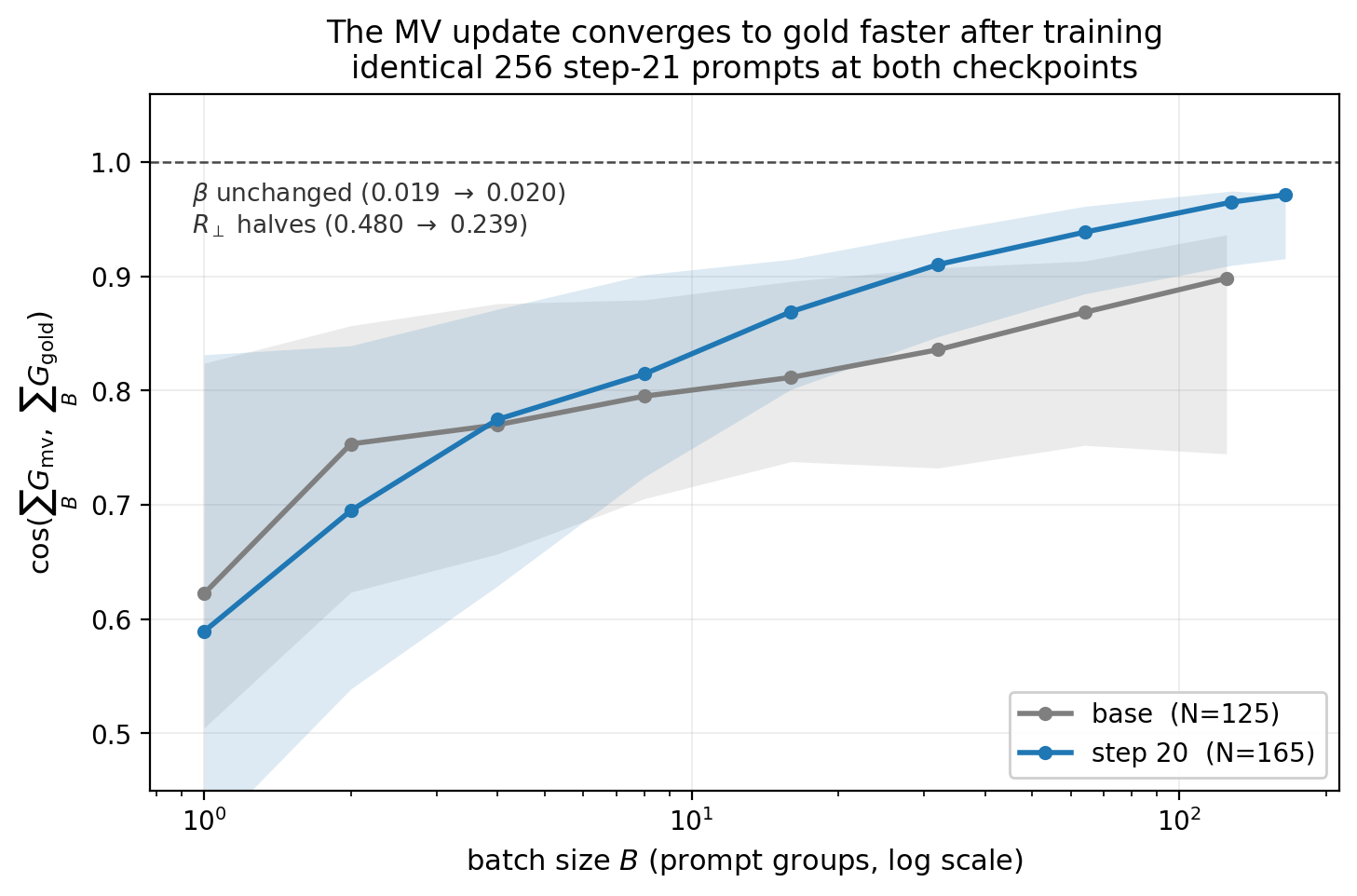}
\caption{The same-prompt comparison of Table~\ref{tab:replication}, as a function of batch size.
Both checkpoints are evaluated on the identical 256 step-21 prompts, each on its own on-policy
rollouts, so the difference between the curves cannot be a difference in prompt composition. Bands
are 95\% bootstrap intervals over prompt groups and overlap at every $B$; the figure illustrates
the decomposition rather than adding evidence for it. The curves cross near $B = 4$: per prompt the
step-20 update is no closer to gold than the base update is, and the gain is entirely a batch
effect, which is what $\beta$ unchanged and $R_\perp$ halved means.}
\label{fig:sameprompt}
\end{figure}

\paragraph{Parsing over training.}
Table~\ref{tab:parserate} tracks answer parsing along the gold run. Parsing improves over the first
eight steps and then declines slowly; it does not break down. The accuracy of the majority answer is
flat across the whole run, so the decline in parse rate is not accompanied by a decline in label
quality. The vote does become steadily more concentrated, which is the same sharpening that
Appendix~\ref{app:additional} identifies as the reason on-policy labels deliver more of the gold
update than frozen ones. The independently seeded MV run of Table~\ref{tab:seedband}, which checkpoints through step 12,
parses at $0.87$ at step 10 and $0.86$ at step 12.

\begin{table}[H]
\centering
\small
\caption{Answer parsing along the gold run of Qwen3-4B-Base, over the 4{,}096 responses generated at
each step. \emph{Parse rate} is the fraction of responses with an extractable answer,
\emph{majority share} the fraction of parsed responses holding the group's plurality answer, and
\emph{maj.\ correct} the fraction of groups whose plurality answer matches the reference.}
\label{tab:parserate}
\begin{tabular}{lccccccc}
\toprule
Step & 1 & 8 & 10 & 15 & 20 & 26 \\
\midrule
Parse rate      & 0.668 & 0.843 & 0.840 & 0.811 & 0.748 & 0.682 \\
Majority share  & 0.382 & 0.408 & 0.420 & 0.465 & 0.511 & 0.548 \\
Maj.\ correct   & 0.344 & 0.348 & 0.398 & 0.352 & 0.316 & 0.367 \\
\bottomrule
\end{tabular}
\end{table}

\paragraph{Evaluation--training overlap.}
We checked all 500 MATH-500 problems against the 29,000 DeepScaleR training problems under three
tests: equality of whitespace- and \LaTeX-normalised strings, substring containment in either
direction, and token-set Jaccard overlap at a 0.80 threshold. One problem matches exactly and 22
exceed the overlap threshold; restricted to the 134 level-5 problems used in
Section~\ref{sec:training}, the counts are one exact and two near matches. The detector was
validated by planting verbatim and reformatted copies of training problems into the evaluation
set, all of which it recovered. Excluding the three flagged level-5 problems changes the
wrong-to-right majority repair rate by less than its bootstrap interval.

\paragraph{Gradient sketches.}
For each response we compute the gradient of $\log\pi_\theta(y\mid x)$, summed over response tokens, at the policy that generated it, and compress it with a count sketch to $K=2^{16}$ dimensions (fixed hash seed). The group gradient under a reward is $\sum_j (r_{ij}-\bar r_i)\, s_{ij}$, where $s_{ij}$ is the sketch of response $j$. A checkpoint from step $t$ is sketched on the responses the run generated at step $t+1$, so the sketches are on-policy. For Qwen3-4B-Base each checkpoint yields $256\times16$ responses. Qwen3-1.7B-Base is sketched on 1536 prompts and the two Llama models on 512 fresh DeepScaleR prompts, with 16 responses each.
Qwen2.5-7B is sketched on 256 prompts with 16 responses each. The pooled Qwen3-4B arm of 512 prompts combines two batches of 256, the second generated after one update, so half of it is off-policy; the on-policy row of Table~\ref{tab:split-half} is the first batch alone, and every result we draw conclusions from is reported for that row as well. Cosines from sketches agree with exact fp32 gradients to within $0.0039$.

\paragraph{Groups with no gold signal.}
Section~\ref{sec:prelim} restricts the comparisons to groups that are mixed under the gold reward,
because a group whose responses all receive the same gold reward contributes the zero vector to
$G_{\mathrm{gold}}$ and admits no cosine with it. MV is frequently still live on the excluded
groups, so the restriction is not neutral for the MV side. Repeating the Qwen3-4B-Base comparison
over all 256 prompts rather than the 164 mixed ones leaves the disattenuated cosine unchanged
($c_\infty = 0.985$, 95\% CI $[0.940, 1.073]$, against $0.999$ $[0.965, 1.043]$) while raising the
norm ratio $\|G_{\mathrm{mv}}\|/\|G_{\mathrm{gold}}\|$ from $1.14$ to $1.62$ and the misdirection
$R_\perp$ from $0.43$ to $0.75$. Including the gold-silent groups therefore makes the MV update
larger without turning it: the extra mass has no gold counterpart to align with, which is what a
group carrying MV signal and no gold signal must contribute.

\paragraph{Split-half estimates.}
$X$, $Y$, and $Z$ are averaged over random partitions of the prompt groups into two halves. Confidence intervals come from a bootstrap over prompt groups in which the groups are partitioned first and then resampled within each half.
Point estimates average $X$, $Y$, and $Z$ over 200 random partitions. Intervals come from 250 bootstrap resamples of the prompt groups, with 12 partitions drawn inside each resample.

\section{Proofs}
\label{app:proofs}

\paragraph{Proposition~\ref{prop:growth}.}
Expanding the squared norm into diagonal and off-diagonal terms gives
$\|\sum_i v_i\|^2 = \sum_i\|v_i\|^2 + \sum_{i\neq j}\langle v_i,v_j\rangle = Bs^2 + \bar\rho\,(B-1)\,Bs^2$.
Treating $s$ and $\bar\rho$ as constant in $B$ and differentiating $\frac12\log\big(Bs^2(1+(B-1)\bar\rho)\big)$ with respect to $\log B$ gives $\alpha(B)$.

\paragraph{Proposition~\ref{prop:misdirection}.}
Prompts whose majority label is correct contribute nothing to $G_{\mathrm{mv}}-G_{\mathrm{gold}}$, so $E_\perp = \sum_{i\in\mathcal W} e_i$, where $\mathcal W$ is the set of $\varepsilon B$ prompts with a wrong majority label. Applying Proposition~\ref{prop:growth} to the $\varepsilon B$ vectors $e_i$ and to the $B$ gold prompt gradients gives
$\|E_\perp\|^2 = \varepsilon B\, s_e^2\,(1+(\varepsilon B-1)\rho_e)$ and $\|G_{\mathrm{gold}}\|^2 = B\, s_g^2\,(1+(B-1)\rho_g)$, and the ratio is $R_\perp^2$. For $\rho_e=0$ the ratio is $\varepsilon (s_e^2/s_g^2)/(1+(B-1)\rho_g)\to 0$. For $\rho_e>0$ and large $B$ it tends to $\varepsilon^2 (s_e^2/s_g^2)(\rho_e/\rho_g)$.

\paragraph{Corollary~\ref{cor:flips}.}
Write the flipped reward as $\tilde r_{ij} = r_{ij} + (1-2r_{ij})\,\zeta_{ij}$ with independent $\zeta_{ij}\sim\mathrm{Bernoulli}(\varepsilon)$. Then $\mathbb{E}[\tilde r_{ij}] = (1-2\varepsilon)\,r_{ij} + \varepsilon$. Mean-centering is linear and removes the constant, so $\mathbb{E}[\tilde A_{ij}] = (1-2\varepsilon)A_{ij}$, and since $g_i$ is linear in the advantages, $\mathbb{E}[g_i(\tilde r)] = (1-2\varepsilon)\,g_i(r)$. Summing over prompts gives $\mathbb{E}[G(\tilde r)] = (1-2\varepsilon)\,G_{\mathrm{gold}}$. The deviations $g_i(\tilde r)-\mathbb{E}[g_i(\tilde r)]$ depend only on the flips of prompt $i$, so they have zero mean and are independent across prompts. The expected update is parallel to the expected gold update, so $c_\infty=1$.

\paragraph{The disattenuated cosine.}
Let the half-batch updates be $G^{\mathcal H}_{k} = m_k + \xi^{\mathcal H}_k$ for $k\in\{\mathrm{mv},\mathrm{gold}\}$ and halves $\mathcal H\in\{\mathcal A,\mathcal B\}$, where $m_k$ is the expected update and the noise from sampling prompts is independent across halves, with the same distribution in both. Because $X$, $Y$, and $Z$ pair different halves, the inner products in their numerators concentrate on $\langle m_{\mathrm{mv}}, m_{\mathrm{gold}}\rangle$, $\|m_{\mathrm{gold}}\|^2$, and $\|m_{\mathrm{mv}}\|^2$, while the noise inflates only the norms in the denominators. The norm factors cancel in $X/\sqrt{YZ}$, which leaves $\langle m_{\mathrm{mv}}, m_{\mathrm{gold}}\rangle/(\|m_{\mathrm{mv}}\|\,\|m_{\mathrm{gold}}\|)$.

\section{Additional Results}
\label{app:additional}

\paragraph{Labels must come from the current policy.}
The labeller's accuracy is again not what matters. At step 20 we relabel the MV run's own responses with the majority answer of the frozen base model on the same prompts. These labels are nearly as accurate as the on-policy majority (a correct majority on 33\% vs.\ 35\% of prompts), yet they deliver only 63\% of the gold component of the update ($\beta = 0.37$), against 98\% for on-policy MV, and add more misdirection ($R_\perp = 0.29$ vs.\ $0.24$). Stale labels shrink the update far more than current ones, in line with the practice of recomputing majority labels from the current policy \citep{zuo2026ttrl,shafayat2025srt}.

\begin{table}[H]
\centering
\small
\caption{Label-free methods recover most of the gain of ground-truth RL. Recovery is $(\text{method}-\text{base})/(\text{GT-RL}-\text{base})$, computed from benchmark-averaged scores reported within the same paper; values are means over settings, with per-setting results in Table~\ref{tab:related-recovery-detail}. For MV, only online variants (labels from the current policy) are included.}
\label{tab:related-recovery}
\begin{tabular}{llccc}
\toprule
Method & Source & Settings & Range & Mean recovery \\
\midrule
\multicolumn{5}{l}{\emph{Online MV}} \\
\quad TTRL$^\dagger$      & \citet{zuo2026ttrl}      & 2 & 109--111\% & 110\% \\
\quad Plurality-vote GRPO & \citet{zhao2026learning} & 2 & \phantom{0}69--98\%\phantom{0}   & \phantom{0}83\% \\
\quad SRT                 & \citet{shafayat2025srt}  & 4 & \phantom{0}55--100\%  & \phantom{0}81\% \\
\midrule
\multicolumn{5}{l}{\emph{Modified label-free rewards}} \\
\quad RESTRAIN            & \citet{yu2026restrain}   & 2 & \phantom{0}81--97\%\phantom{0}   & \phantom{0}89\% \\
\quad INTUITOR            & \citet{zhao2026learning} & 6 & \phantom{0}76--111\%  & \phantom{0}96\% \\
\bottomrule
\end{tabular}
\par\vspace{3pt}
\footnotesize
\raggedright
$^\dagger$TTRL trains on evaluation questions, whereas its ground-truth reference (Oat-Zero) trains on MATH-train; this row therefore overstates recovery relative to the others.
\end{table}

\begin{table}[H]
\centering
\small
\caption{Errors grow like a random walk while the gold signal grows coherently. Local growth exponents $\alpha(B)$ of summed gradients between successive batch sizes, Qwen3-4B-Base. By Proposition~\ref{prop:growth}, $\alpha = 1/2$ indicates incoherent vectors and $\alpha \to 1$ a dominant shared direction.}
\label{tab:exponents}
\begin{tabular}{llccc}
\toprule
 & & \multicolumn{2}{c}{Growth exponent $\alpha(B)$} & \\
\cmidrule(lr){3-4}
Checkpoint & $B$ range & MV error $E$ & Gold & Gap \\
\midrule
Base    & \phantom{00}8$\,\to\,$16  & 0.524 & 0.581 & 0.057 \\
        & \phantom{0}32$\,\to\,$64  & 0.500 & 0.720 & 0.220 \\
        & \phantom{0}64$\,\to\,$128 & 0.524 & 0.813 & 0.289 \\
        & 128$\,\to\,$164           & 0.509 & 0.874 & 0.365 \\
\midrule
Step 20 & \phantom{0}16$\,\to\,$32  & 0.506 & 0.684 & 0.179 \\
        & \phantom{0}64$\,\to\,$128 & 0.503 & 0.835 & 0.333 \\
        & 128$\,\to\,$165           & 0.486 & 0.901 & 0.415 \\
\bottomrule
\end{tabular}
\end{table}

\begin{table}[H]
\centering
\small
\setlength{\tabcolsep}{5pt}
\caption{Full split-half results behind Table~\ref{tab:split-half}. $\Delta$ is the 95\% interval of (MV vs.\ gold) minus (gold vs.\ gold); an interval containing zero means MV cannot be distinguished from gold. MV vs.\ MV is the agreement of MV with itself across halves.}
\label{tab:split-half-full}
\begin{tabular}{lrcccc}
\toprule
Model & $N$ & Gold vs.\ gold & MV vs.\ gold & MV vs.\ MV & $\Delta$ (95\% CI) \\
\midrule
Qwen3-4B-Base$^\dagger$  & 331 & 0.647 & 0.647 & 0.646 & \ci{$-$.014}{$+$.014} \\
Qwen3-1.7B-Base          & 636 & 0.842 & 0.834 & 0.831 & \ci{$-$.016}{$-$.003} \\
Qwen2.5-7B               & 151 & 0.799 & 0.818 & 0.843 & \ci{$+$.010}{$+$.044} \\
Llama-3.2-3B-Instruct    & 247 & 0.371 & 0.360 & 0.352 & \ci{$-$.022}{$+$.010} \\
Llama-3.1-8B-Instruct    & 208 & 0.438 & 0.399 & 0.362 & \ci{$-$.050}{$-$.016} \\
\bottomrule
\end{tabular}
\par\vspace{3pt}
\footnotesize
\raggedright
$^\dagger$Pools two batches of 256 prompts; the second was generated after one update.
\end{table}

\begin{table}[H]
\centering
\small
\setlength{\tabcolsep}{5pt}
\caption{Full version of Table~\ref{tab:coherence}, adding the MV gradients themselves. Mean pairwise cosine between prompt gradients ($\times10^{-3}$), Qwen3-4B-Base along the MV run.}
\label{tab:coherence-full}
\begin{tabular}{lccccccccc}
\toprule
 & \multicolumn{4}{c}{Right majority} & & \multicolumn{4}{c}{Wrong majority} \\
\cmidrule(lr){2-5} \cmidrule(lr){7-10}
Checkpoint & Gold & MV & MV error $E$ & Null & & Gold & MV & MV error $E$ & Null \\
\midrule
Base    & 22.41 & 20.06 & 0.39 & 0.30 & & \phantom{0}6.55 & \phantom{0}9.26 & 0.23 & 0.69 \\
Step 10 & \phantom{0}6.52 & \phantom{0}6.52 & 0.01 & 0.36 & & \phantom{0}0.82 & \phantom{0}2.79 & 0.15 & 0.50 \\
Step 20 & 27.99 & 26.94 & 0.14 & $-$0.15 & & 13.44 & 14.71 & 0.55 & 0.66 \\
\bottomrule
\end{tabular}
\end{table}

\begin{table}[H]
\centering
\small
\caption{MV training repairs wrong majorities on held-out problems. 134 level-5 MATH-500 problems, base model against step 20 of the MV run (66 wrong-majority and 68 right-majority problems at base); majorities use all 64 samples. (a) maj@$k$ and its change, with a 95\% bootstrap interval over problems. (b) Problems whose majority answer changes; the resampling null is the flip rate between two disjoint halves of 32 samples at step 20. $^{*}$The interval excludes zero in (a) and the resampling null in (b).}
\label{tab:maj-flips}
\begin{tabular}{lccc}
\toprule
\multicolumn{4}{l}{\emph{(a) Majority-vote accuracy}} \\
 & Base & Step 20 & Change (95\% CI) \\
\midrule
maj@1  & 0.185 & 0.374 & $+$0.188 \ci{$+$.158}{$+$.218}$^{*}$ \\
maj@4  & 0.292 & 0.466 & $+$0.174 \ci{$+$.145}{$+$.206}$^{*}$ \\
maj@16 & 0.443 & 0.543 & $+$0.100 \ci{$+$.066}{$+$.136}$^{*}$ \\
maj@64 & 0.503 & 0.585 & $+$0.082 \ci{$+$.030}{$+$.138}$^{*}$ \\
\midrule
\multicolumn{4}{l}{\emph{(b) Majority-answer flips, base $\rightarrow$ step 20}} \\
 & Observed & Rate, \% (95\% CI) & Resampling null \\
\midrule
Wrong $\rightarrow$ right & 13/66 & 19.7 \ci{10.6}{28.8}$^{*}$ & 4/55 (7.3\%) \\
Right $\rightarrow$ wrong & \phantom{0}3/68  & \phantom{0}4.4 \ci{0.0}{9.3} & 10/79 (12.7\%) \\
\bottomrule
\end{tabular}
\end{table}

\begin{figure}[H]
\centering
\includegraphics[width=0.6\linewidth]{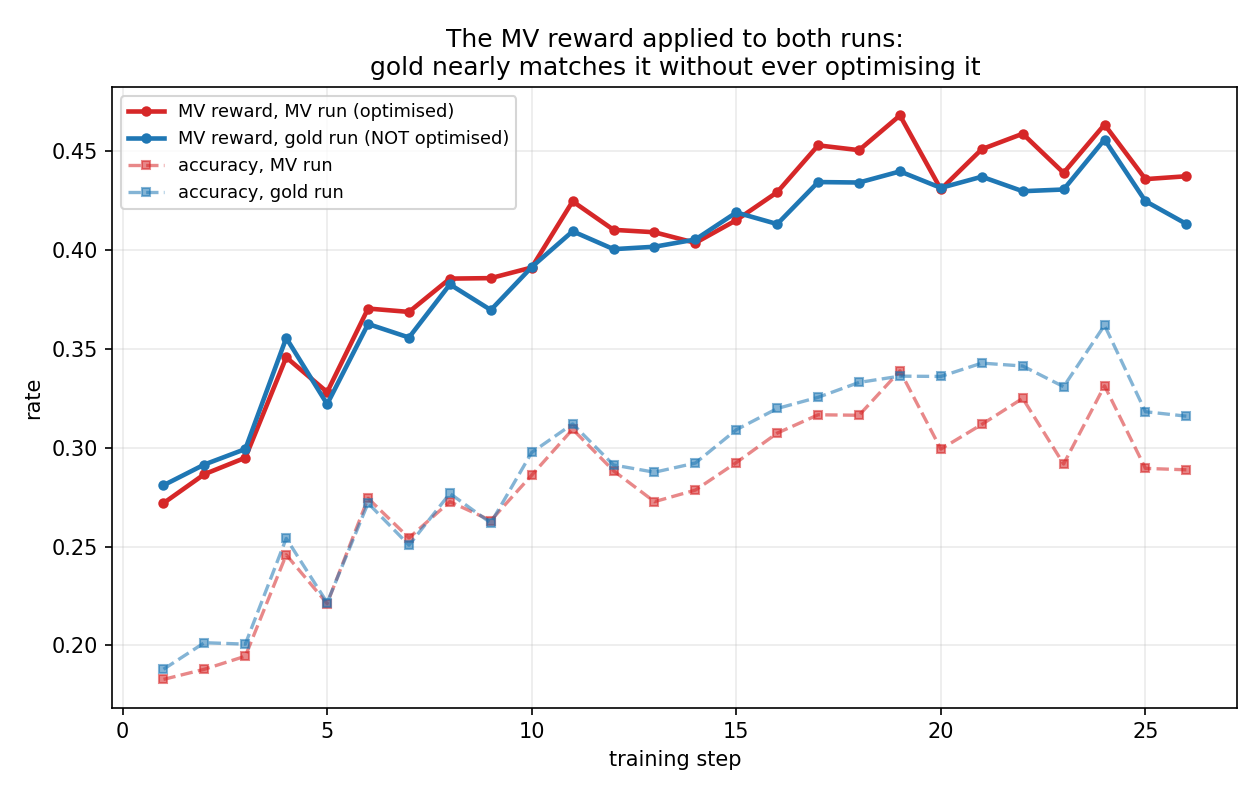}
\caption{MV reward (solid) and accuracy (dashed) on the training prompts of the gold and MV runs of Qwen3-4B-Base. The gold run nearly matches the MV run on the MV reward without ever optimizing it.}
\label{fig:mv-reward-gold}
\end{figure}

\begin{table}[H]
\centering
\small
\caption{INTUITOR's alignment with gold depends on the model. Disattenuated cosine $c_\infty$ between the INTUITOR and gold updates at initialization, raw and after controlling for response length.}
\label{tab:intuitor}
\begin{tabular}{lcc}
\toprule
Model & Raw & Length-controlled (95\% CI) \\
\midrule
Qwen3-4B-Base$^\dagger$   & 0.990 & 0.973 \ci{.934}{1.029} \\
Qwen3-4B-Base (on-policy) & 0.950 & 0.941 \ci{.871}{.993} \\
Qwen3-1.7B-Base           & 0.902 & 0.969 \ci{.955}{.971} \\
Qwen2.5-7B                & 0.536 & 0.959 \ci{.926}{1.017} \\
Llama-3.2-3B-Instruct     & 0.855 & 0.786 \ci{.727}{.862} \\
Llama-3.1-8B-Instruct     & 0.883 & 0.756 \ci{.738}{.825} \\
\bottomrule
\end{tabular}
\par\vspace{3pt}
\footnotesize
\raggedright
$^\dagger$The pooled arm of Table~\ref{tab:split-half}.
\end{table}

\begin{table}[H]
\centering
\caption{Per-setting numbers behind Table~\ref{tab:related-recovery}. Scores are the average over the listed benchmarks (pass@1, \%), including GSM8K \citep{cobbe2021training}.
Recovery $=(\text{Method}-\text{Base})/(\text{GT-RL}-\text{Base})$. SRT values are read off \citet{shafayat2025srt}, Fig.~3 (final step), and are approximate.}
\label{tab:related-recovery-detail}
\small
\resizebox{\linewidth}{!}{%
\begin{tabular}{llllrrrr}
\toprule
Method & Model & Train set & Benchmarks averaged & Base & GT-RL & Method & Recovery \\
\midrule
\multicolumn{8}{l}{\emph{\citet{zuo2026ttrl}} --- GT-RL = Oat-Zero (same backbone), greedy decoding} \\
TTRL & Qwen2.5-Math-1.5B & test set & AIME24, AMC, MATH-500 & 28.5 & 49.1 & 51.0 & 109\% \\
TTRL & Qwen2.5-Math-7B   & test set & AIME24, AMC, MATH-500 & 35.3 & 62.0 & 65.0 & 111\% \\
\midrule
\multicolumn{8}{l}{\emph{\citet{zhao2026learning}} --- GT-RL = GRPO with gold answers, greedy decoding} \\
Plurality-vote GRPO & Qwen2.5-3B        & MATH & GSM8K, MATH-500 & 60.9 & 73.1 & 72.8 & 98\% \\
Plurality-vote GRPO & Llama-3.2-3B-Inst & MATH & GSM8K, MATH-500 & 56.2 & 60.4 & 59.1 & 69\% \\
INTUITOR & Qwen2.5-1.5B      & MATH & GSM8K, MATH-500 & 4.6  & 65.4 & 62.1 & 95\% \\
INTUITOR & Qwen2.5-3B        & MATH & GSM8K, MATH-500 & 60.9 & 73.1 & 70.2 & 76\% \\
INTUITOR & Qwen2.5-7B        & MATH & GSM8K, MATH-500 & 59.5 & 79.0 & 81.2 & 111\% \\
INTUITOR & Qwen2.5-14B       & MATH & GSM8K, MATH-500 & 71.3 & 83.8 & 84.7 & 107\% \\
INTUITOR & Llama-3.2-3B-Inst & MATH & GSM8K, MATH-500 & 56.2 & 60.4 & 60.0 & 90\% \\
INTUITOR & OLMo-2-7B-SFT     & MATH & GSM8K, MATH-500 & 49.7 & 54.2 & 54.1 & 98\% \\
\midrule
\multicolumn{8}{l}{\emph{\citet{shafayat2025srt}} --- GT-RL = RLOO with gold answers} \\
SRT & Llama-3.1-8B-Inst     & Big-Math & MATH-500 & 52.7 & 59.0 & 58.5 & 92\% \\
SRT & Qwen2.5-Math-7B       & MATH-12K & MATH-500 & 43   & 78   & 78.5 & 100\% \\
SRT & Qwen3-14B-Base        & DAPO     & MATH-500 & 52   & 86.5 & 71   & 55\% \\
SRT & DeepSeek-Math-7B-Inst & MATH-12K & MATH-500 & 43.0 & 47.5 & 46.4 & 76\% \\
\midrule
\multicolumn{8}{l}{\emph{\citet{yu2026restrain}} --- GT-RL = GRPO with gold labels; 16 samples, $T{=}0.6$} \\
RESTRAIN & Qwen3-4B-Base  & DAPO-14k & 6 benchmarks$^\ast$ & 38.0 & 51.4 & 51.0 & 97\% \\
RESTRAIN & OctoThinker-8B & DAPO-14k & 6 benchmarks$^\ast$ & 19.2 & 39.6 & 35.8 & 81\% \\
\bottomrule
\end{tabular}%
}

\par\vspace{3pt}
\footnotesize
\raggedright
$^\ast$MATH-500, AIME25, OlympiadBench \citep{he-etal-2024-olympiadbench}, Minerva \citep{lewkowycz2022solving}, MMLU-STEM, GPQA-Diamond. OctoThinker is from \citet{wang2025octothinker}. On Qwen3-4B, RESTRAIN exceeds GT-RL on MMLU-STEM and GPQA-Diamond; on the four math benchmarks alone its recovery is about 80\%.
\end{table}

\begin{table}[H]
\centering
\small
\caption{The per-step MV errors do not share a direction (Section~\ref{sec:temporal}).
Qwen3-4B-Base along the MV run, each checkpoint sketched on its own on-policy rollouts.
\emph{Coherence} is the mean cosine over the ten step pairs $t \neq s$; the random reward is
evaluated through the same pipeline and its expected update is the zero vector, so it gives the
floor. \emph{Exponent} is the log--log slope of $\|\sum_{t\le T} v_t\|$ in $T$. The sign-flip
reference randomises the sign of each $E_t$, preserving all norms while destroying any consistent
direction, and so is the $\sqrt{T}$ null for these data.}
\label{tab:temporal}
\begin{tabular}{lcc}
\toprule
 & Coherence $\cos(v_t, v_s)$ & Growth exponent \\
\midrule
MV error $E_t$            & 0.030 & 0.594 \\
Gold gradient             & 0.400 & 0.762 \\
Random reward (floor)     & 0.001 & -- \\
\midrule
Sign-flip reference ($\sqrt{T}$ null) & -- & 0.540 \ci{.522}{.576} \\
\bottomrule
\end{tabular}
\par\vspace{3pt}
\footnotesize
\raggedright
A single sketched cosine resolves $0.004$. Steps $t=3,\dots,11$ (every checkpoint the run saved); $N=163$--$174$ mixed-gold groups per step.
The observed exponent exceeds the sign-flip reference with $p = 0.005$.
\end{table}

\begin{table}[H]
\centering
\small
\caption{A prompt's own gradient barely outweighs another prompt's. Change in the output distribution of prompt $q$ after a small step along $q$'s own gradient ($d_\text{own}$) or along another prompt's gradient ($d_\text{other}$), measured on identical tokens (Qwen3-1.7B-Base, 8 prompts, fp64). A fully tabular model would give own/other $\to\infty$; a fully shared one would give 1.}
\label{tab:locality}
\begin{tabular}{lccc}
\toprule
Step size & $d_\text{own}$ & $d_\text{other}$ & Own/other \\
\midrule
$10^{-7}$ & $7.18\times10^{-6}$ & $3.70\times10^{-6}$ & 1.94 \\
$10^{-8}$ & $7.23\times10^{-8}$ & $4.19\times10^{-8}$ & 1.72 \\
\bottomrule
\end{tabular}
\end{table}

\begin{table}[H]
\centering
\small
\setlength{\tabcolsep}{5pt}
\caption{A single prompt's contribution shrinks with batch size. \emph{Own share}: $d_\text{own}/d_\text{batch}$, the fraction of the change in a prompt's output caused by its own gradient within a batch of size $B$ (step size $10^{-7}$). \emph{Scaling}, computed over every prompt group from the gradient sketches: the KL-share slope is the log--log slope of the KL change from swapping one prompt against $B$ (i.i.d.\ theory: $-2$); writing $g_q=\mu+\xi_q$, the shared energy is $\|\mu\|^2$ as a fraction of the total, and $B^\ast$ is the batch size at which the shared component overtakes the prompt-specific one.}
\label{tab:prompt-share}
\begin{tabular}{lccccccc}
\toprule
 & \multicolumn{4}{c}{Own share (\%)} & \multicolumn{3}{c}{Scaling} \\
\cmidrule(lr){2-5} \cmidrule(lr){6-8}
Model & $B{=}8$ & $B{=}16$ & $B{=}64$ & $B{=}128$ & KL-share slope & Shared energy & $B^\ast$ \\
\midrule
Qwen3-1.7B-Base       & \phantom{0}9.81 & \phantom{0}3.37 & 0.38 & 0.13 & $-$1.37 & 2.27\% & 43 \\
Qwen3-4B-Base         & \phantom{0}3.36 & \phantom{0}0.86 & 0.18 & 0.09 & $-$1.31 & 2.63\% & 37 \\
Llama-3.2-3B-Instruct & 29.74 & 25.69 & 0.83 & --   & $-$1.15 & 1.49\% & 66 \\
\bottomrule
\end{tabular}
\par\vspace{3pt}
\footnotesize
\raggedright
For Llama-3.2-3B-Instruct the share barely falls from $B=8$ to $16$ because the small-batch gradient is noisy ($d_\text{batch}$ moves only from $1.63\times10^{-5}$ to $1.89\times10^{-5}$).
\end{table}

\end{document}